%% file: main.tex
\pdfoutput=1

\documentclass[11pt]{article}

\usepackage{acl}

\usepackage{times}
\usepackage{latexsym}

\usepackage[T1]{fontenc}
\usepackage[utf8]{inputenc}

\usepackage{microtype}

\usepackage{times,latexsym}
\usepackage{url}
\usepackage[T1]{fontenc}
\usepackage{graphicx}
\usepackage{amsmath}
\usepackage{amsthm}
\usepackage{amssymb}

\usepackage[shortlabels]{enumitem}
\usepackage[most]{tcolorbox}
\usepackage{makecell}
\usepackage{booktabs}
\usepackage{enumitem}
\usepackage{cleveref}
\usepackage{color,soul}
\usepackage{tabularx,colortbl}
\usepackage{multirow}
\usepackage{tablefootnote}

\usepackage{xspace}

\usepackage{tasks}

\newcommand{\ie}{i.e.,~}
\newcommand{\eg}{e.g.,~}

\newcommand{\sqhints}{\textsc{Single-Hints}\xspace}
\newcommand{\mqhints}{\textsc{Multi-Hints}\xspace}

\newcommand{\rsb}{\textsc{RSB}\xspace}
\newcommand{\tb}{\textsc{TB}\xspace}
\newcommand{\dhg}{\textsc{DHG}\xspace}
\newcommand{\ptg}{\textsc{PTG}\xspace}
\newcommand{\hintstat}{\textsc{$6,681$} }

\usepackage[htt]{hyphenat}

\newcommand{\example}[1]{{``\emph{#1}''}\normalsize}
\newcommand{\domain}[1]{{\texttt{#1}}\normalsize}

\title{Follow-on Question Suggestion via Voice Hints for Voice Assistants}

\author{Besnik Fetahu$^1$, Pedro Faustini$^{2}$\thanks{~~Work done during an internship at Amazon.}, Giuseppe Castellucci$^1$, Anjie Fang$^1$ \\\textbf{Oleg Rokhlenko$^1$, Shervin Malmasi$^1$} \\
 Amazon, Seattle, WA, USA \\
 Macquarie University, Sydney, NSW, Australia \\
\texttt{pedro.arrudafaustini@hdr.mq.edu.au} \\ \texttt{\{besnikf,giusecas,njfn,olegro,malmasi\}@amazon.com}}

\newcommand{\change}[1]{\textcolor{black}{#1}}

\date{}

\begin{document}
\maketitle
\begin{abstract}

The adoption of voice assistants like Alexa or Siri has grown rapidly, allowing users to instantly access information via voice search. 
Query suggestion is a standard feature of screen-based search experiences, allowing users to explore additional topics. However, this is not trivial to implement in voice-based settings.
To enable this, we tackle the novel task of suggesting questions with compact and natural \emph{voice hints} to allow users to ask follow-up questions.
We define the task, ground it in syntactic theory and outline linguistic desiderata for spoken hints.
We propose baselines and an approach using sequence-to-sequence Transformers to generate spoken hints from a list of questions.
Using a new dataset of \hintstat input questions and human written hints, we evaluated the models with automatic metrics and human evaluation.
Results show that a naive approach of concatenating suggested questions creates poor voice hints. Our approach, which applies a linguistically-motivated pretraining task was strongly preferred by humans for producing the most natural hints.

\end{abstract}

\input{introduction}
\input{task}

\input{approach}
\input{dataset}

\input{setup}

\input{evaluation}

\input{related_work}

\input{conclusions}

\section*{Limitations}
\label{sec:limitations}

\paragraph{\textbf{Languages.}} We limited our work to the English language for obtaining training and testing data for generating voice-friendly hints. As a next step, we foresee adding other languages, such as German, Korean, and Chinese, and understanding the implications in terms of the required syntactic and semantic operations to generate voice-friendly hints.

\paragraph{\textbf{Scenarios.}} Our work focused only on a single turn conversations, where after a user asks a question to a voice assistant, a hint suggesting related questions are uttered back to the user. Future steps include multi-turn conversations, where user interests and actions after each hint will impact the generated hints for follow-up turns. There are several strategies that can be considered, and we aim at investigating the following: dive deeper in a topic of user's interest (suggest more targeted questions on a specific topic about the entity of interest), or broaden user's knowledge on a given topic (i.e., suggest questions about \emph{related entities}).

\paragraph{\textbf{Large Language Models.}} While in this work we do not focus on recent multi-billion parameter LLMs, in \Cref{sec:llm_hints} we present an evaluation of the performance of ChatGPT on our test set for the task of generating voice friendly hints. We do not go in depth in our analysis for ChatGPT and similarly large models for two key reasons. First, ChatGPT can be considered as a black box, where there is no scientific reporting on the models parameters and its training. Second, due to the strict latency requirements in voice assistants, such large models are not feasible to be used for applications like ours where the hint must be generated in 150 milliseconds or less.

\bibliographystyle{acl_natbib}
\bibliography{main}

\appendix
\include{appendix_v2}
\end{document}

%% file: introduction.tex
\section{Introduction}\label{sec:introduction}

Voice assistants, like Alexa
or Google Assistant
provide ubiquitous services through a variety of devices (e.g. smart speakers, phones, TVs, etc.). Users interact with voice assistants for different purposes~\cite{rzepka2019examining,lopatovska2019talk} such as question answering, e-commerce, or entertainment. With increasing adoption, user expectations also grow and related content recommendation is a valued feature~\cite{DBLP:journals/imwut/TabassumKFMWEL19}.
\begin{figure}
    \centering
    \includegraphics[width=1\columnwidth]{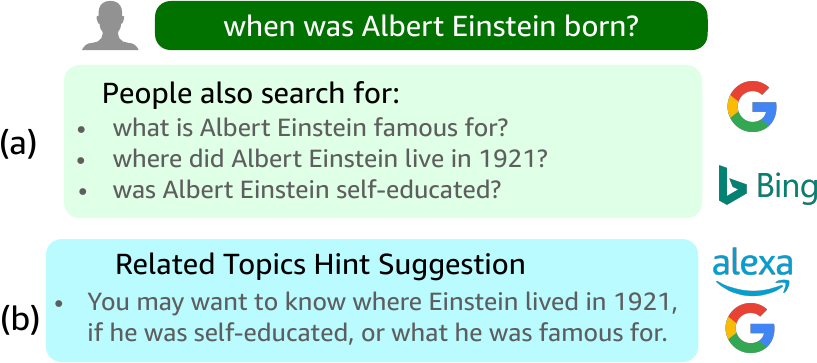}
    \caption{{(a) Question suggestion in web search (available in Google/Bing) for a user question. (b) Proposed voice-based hint for the same questions users can ask as follow-on questions to a voice assistant such as Alexa.}} 
    \label{fig:voice_hint_example}
\end{figure}

The question of \textit{how} to present proactive suggestions is an open one, and recent work has examined how content such as news articles can be recommended over voice \cite{10.1145/3343413.3378013}.
Query and question recommendation (see Fig.~\ref{fig:voice_hint_example} (a)) have become well-established research topics, and are well integrated in screen-based Web search experiences (\ie those from Google/Bing). However, such functionality does not exist for voice-based systems.
Suggestions enable highly useful exploratory search capabilities, and we aim to provide a similar experience over voice (see Fig.~\ref{fig:voice_hint_example} (b)), where through a \emph{follow-on hint} we suggest related topics they can ask about.

Contrary to suggestions on Web search, integrating recommendations in voice assistants poses unique challenges~\cite{DBLP:conf/cui/MaL20}, such as (i) \emph{modality}: voice lacks the advantages of visual interfaces used on the Web (\eg showing a list), (ii) \emph{transmitted information}: to ensure comprehension, the amount of transmitted information in an utterance is limited in terms of time and number of words, and (iii) \emph{shape}: simply reading out a list of questions is not natural over voice. 

We propose a new approach on \textbf{\emph{how}} to deliver voice-based question suggestions using hints. We do not consider \textbf{\emph{what}} to suggest as this is widely explored in existing work.
Figure~\ref{fig:voice_hint_example} provides an overview. For an input question, we assume the voice assistants can retrieve related questions\footnote{Related questions can be log based, or retrieved from a question bank using a similarity metric.} from which a suggestion hint is generated. Differently from questions recommendation in Web search (Fig.~\ref{fig:voice_hint_example} (a)) where new related questions are listed, we aim to synthesize a natural utterance (Fig.~\ref{fig:voice_hint_example} (b)) suggesting the same questions.

The hint does not contain questions, rather, it contains several subordinate clauses describing facts or knowledge that the user can ask about.

Our overarching contribution is a framework for generating voice-friendly hints. We begin with a grounded linguistic description of the task, outlining the characteristics of a good hint (\eg cohesion, length), and the syntactic transformations needed to construct such utterances. Next, we frame the task as a \emph{seq2seq} approach~\cite{DBLP:conf/acl/LewisLGGMLSZ20}, where for an input question and its top--3 related questions, covering a diverse set of topics (unrelated topics to the initial question's topic), a voice hint is synthesized to meet the desiderata in \Cref{tab:hint_requirements}.
\change{While newer large language models like ChatGPT are very capable in tasks like ours \cite{ouyang2022training}, generating real-time voice hints requires low latency (<150ms), which cannot be met by such models, hence the need for our task-specific model.}

We create a dataset of voice-friendly hints, consisting of the triple: \emph{initial question}, \emph{related questions}, \emph{follow-on hint}, in 9 different domains.
We evaluate hint generation on our dataset by means of automated metrics and human evaluation studies. %
To summarize, our contributions are:
\begin{enumerate}
    \item To our knowledge, we are the first to define the task of question suggestion via voice hints;
    \item A large real-world hint generation dataset of \hintstat instances, covering 9 domains, that will become publicly available;\footnote{\tiny{\url{https://github.com/bfetahu/spoken_hints/}}}
    \item A seq2seq approach with task-specific training strategies for voice hint generation;
    \item A detailed human evaluation protocol for evaluating different aspects of voice hints.
\end{enumerate}

%% file: task.tex
\section{Linguistic Task and Background}
\label{sec:linguistic_task}

\begin{figure*}[!h]
    \centering
    \includegraphics[width=.9\textwidth]{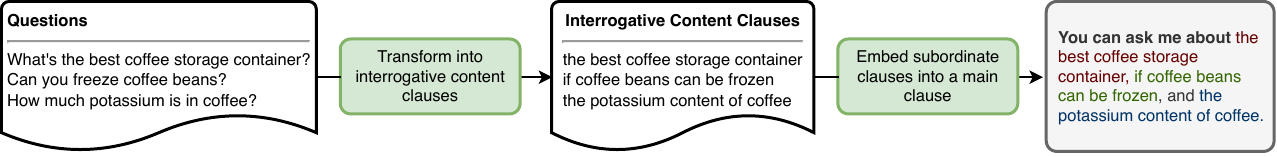}
    \caption{\small{An overview of the linguistic processes for transforming a set of questions into a declarative statement.}}
    \label{fig:linguistic_task}
\end{figure*}

\change{To generate a spoken hint, our objective is to take a set of standalone questions (interrogative sentences), and convert them into a single sentence that informs the listener about the different pieces of information available.} Figure~\ref{fig:linguistic_task} shows the overview of the linguistic tasks that are needed to be performed in order for a set of input questions to generate a voice-friendly hint.

Direct questions (\example{can a dog eat peanuts?}) can be presented as an \emph{indirect question} (\example{Alice asked if dogs can eat peanuts.}) \cite{suner1993indirect}. All direct questions can have an indirect equivalent, and the embedded clause of the indirect version is said to refer to the direct question \cite{puigdollers1999indirect}. 

While both direct and indirect questions can be used to \textit{ask}, when an indirect question's main clause reports information (e.g. \example{I know $\ldots$}), their pragmatic purpose is to \textit{provide} information \cite{puigdollers1999indirect}.
Our task requires transforming independent questions into \emph{subordinate clauses}, and then embedding them into a new sentence whose main verb is one of cognition or reporting, and takes the clauses as direct objects (\Cref{a_subsec:linguistic_task}).

\change{The most interesting syntactic transformation is that of converting a question to a dependent clause. In English, this can be done using content clauses (also known as noun clauses), which describe the inquired information in a main clause. The contents of a question can be framed as an \textit{interrogative content clause} which represents the knowledge or entity that is being interrogated in the question.}

The syntactic transformations needed to construct the content clause vary depending on the question type and its complexity. In general, these are the same changes used to generate reported or indirect speech, and can include subject-auxiliary inversion, changes in tense, and other lexical substitutions. 
This resulting subordinate clause is a syntactic unit which can be used as a direct object in a declarative sentence. Multiple subordinates can be combined to compose a single sentence.

Since these transformations between direct and reported speech are commonly used in English, representing our questions this way sounds very natural, and allows listeners to effortlessly convert any of the clauses into a fully formed question.

\subsection{Characteristics of Natural Voice Hints}

\begin{table*}[!htb]

    \centering
    \resizebox{.85\textwidth}{!}{
    \begin{tabular}{p{2cm}  p{15cm}}
    \toprule
    \textbf{\emph{Aspect}} & \textbf{\emph{Description}}\\
    \midrule
    \emph{Naturalness} & The hint should reference facts or knowledge that can be asked.\\
    \hline
    \emph{Actionability} & The main hint clause should be action oriented, 
    \eg \emph{\textcolor{darkblue}{You} \textcolor{red}{can/may/might/could} \textcolor{olive}{ask/also ask/be interested/also be interested}}.\\
    \hline
    \emph{Information content} & Questions must be converted to an interrogative content clause, just as they would be embedded in an indirect version of the same question.\\
    \hline
    \emph{Length} & The hint utterance should not be exceedingly long in terms of words and listening time.\\
    \hline
    
    \emph{Coherence, ~ Cohesion} &
        \textbullet~The hint is syntactically \change{correct} and semantically coherent.\newline
        \textbullet~Subordinate clauses $Q_{rel}$ are connected through \emph{coordinating conjunctions}~\cite{DBLP:journals/coling/WebberSJK03}.\newline
        \textbullet~Lexical repetitions, e.g., entity mentions, should be replaced by anaphora where appropriate.
        \\
    \bottomrule
    \end{tabular}}
    \caption{Linguistic properties of a natural spoken hint.}
    \label{tab:hint_requirements}
\end{table*}

For a hint to be considered voice-friendly, \ie sound like a natural spoken utterance, several aspects detailed in \Cref{tab:hint_requirements} must be fulfilled.

These desiderata are based on the principles of cohesion and coherence \cite{Halliday76a} and Gricean maxims of conversation \cite{grice1975logic}. They ensure that constructed hints sound natural and are easy to comprehend.
The characteristics were derived from our preliminary experiments on how English speakers create hints.

\subsection{Voice-Friendly Hint Generation Task}
\label{sec:task}

The task is orchestrated as follows: (a) for an input question $q$, defined as a sequence of tokens ${q}=\{x_1, \ldots, x_n\}$ with a subject entity $e$;
(b) a \emph{follow-on hint} is generated from a set of top--$k$ questions ${Q}_{rel}$ about $e$ (cf. \S\ref{a_subsec:preprocessing}), which cover \emph{related topics} not covered in $q$. The generated hint does not contain explicit questions, but related topics that the user can ask about $e$.

The task is to learn the mapping function $\mathcal{F}\left({q}, {Q}_{rel}\right) \rightarrow h$, which learns the transformation described in \S\ref{sec:linguistic_task}, i.e., mapping $q$ and $Q_{rel}$ into $h$, and meets the criteria in \Cref{tab:hint_requirements}, with the most challenging tasks being:

\paragraph{\textbf{Content Clause Generation:}} $\mathcal{F}$ must map $Q_{rel}$ into subordinate content clauses in reported speech format~\cite{lucy1993reflexive}, e.g. \example{how many children does Cristiano Ronaldo have?} $\rightarrow$ \example{Alice asked about \underline{how many children Cristiano Ronaldo has}}.\footnote{Pronoun and verb tense changes are required.}

\paragraph{\textbf{Anaphora:}} ${q}_{rel}\in {Q}_{rel}$ typically contain variable surface forms of $e$, hence its repetitions in $h$ are unnatural. $\mathcal{F}$ needs to learn how and when to replace $e$ in $q_{rel}$ with anaphoric expressions.

%% file: approach.tex
\section{Method}
\label{sec:methodology}

\subsection{Hints Generation Architecture}\label{subsec:hint_generation}

The function $\mathcal{F}(q, Q_{rel})$ corresponds to a generative Transformer model~\cite{Transformers}, which for an input question $q$ and its top--$k$ related questions $Q_{rel}$ produces the hint $h$. We experiment with BART~\cite{DBLP:conf/acl/LewisLGGMLSZ20} and T5 models~\cite{2020t5}.

We encode the input question $q$ and its related questions $Q_{rel}$ as follows:
{$$\mathbf{s} := \{q, \text{\texttt{[SEP]}}, q^1_{rel}, \text{\texttt{[SEP]}}, q^2_{rel}, \text{\texttt{[SEP]}}, q^3_{rel}\}$$}

Representation $\mathbf{s}$ is used by the decoder to generate the hint $h$. 
During the training of $\mathcal{F}$, the model learns to map the input $\mathbf{s}$ to $h$ through operations, such as: (i) using \emph{start patterns}, serving as the main clause of $h$, (ii) converting $Q_{rel}$ into subordinate clauses, (iii) avoid entity repetitions through \emph{anaphora}, and, (iv) ensuring hint \emph{coherence} by connecting the subordinate clauses.

While seq2seq models show remarkable natural language generation capabilities, fine-tuning them for all the criteria above is challenging, resulting in hints that are \emph{incoherent} and \emph{unnatural} (cf. \S \ref{sec:evaluation}). Hence, we propose a pretraining strategy to overcome such challenges.

\subsection{Reported Speech Pretraining}
\label{subsec:rs_pretraining}

A key aspect of ensuring that $h$ is correct is creating the subordinate clauses from $Q_{rel}$, as they would be in reported speech (RS) format. Generating RS requires the model $\mathcal{F}$ to perform the most significant rewrite operations, including performing the subordinate clause syntax change, such as verb tense, pronoun and word order alterations. 

\change{We propose a two stage training strategy, where: (1) we pretrain $\mathcal{F}$ in converting individual questions into their RS format, and finally (2) fine-tune $\mathcal{F}$  for the full hint generation task, ensuring that the hint is coherent and there are no repetitions.}

\noindent\textbf{RS Pre-training:} \change{For pretraining, we change the input of the model to be a single question and output its reported speech equivalent. This is the same as generating a hint from a single question, with the only difference that there is no initial input question $q$ to the model. }
\change{Constraining the pretraining phase to a single question it allows the model to learn how to perform all the necessary rewrite operations for converting a question to RS format.}

\noindent\textbf{{Fine-Tuning:}} \change{Next, we fine-tune the pretrained model to learn to convert the input $\mathbf{s}$ (containing the  $q$ and its related questions $Q_{rel}$) into a hint.}
By this stage, the model already has pretrained knowledge for converting questions into RS, and can focus on learning to use anaphora, conjunctions, etc.

%% file: dataset.tex
\section{VoFH -- Voice-Friendly Question Suggestion via Hints Dataset}\label{sec:dataset}

We now describe the process of generating a new voice-friendly hints dataset.\footnote{{\url{https://github.com/bfetahu/spoken_hints/}}}
We first construct tuples of input questions and related questions $\langle q, {Q}_{rel}\rangle$.
We then annotate spoken hints for each tuple, creating a dataset of \hintstat samples composed of the triples ${Q}=\{\langle q, {Q}_{rel}, h\rangle_i \ldots\}$. The input question $Q$ and related questions $Q_{rel}$ datasets are described in Appendix~\ref{a_subsec:question_dataset}.

\subsection{Hint Annotation}
Using the question bank $Q$ and the related questions $Q_{rel}$, we collect spoken hints for suggesting related questions. From a random sample of \hintstat input questions and their related questions $Q_{rel}$, we create two \emph{disjoint} hint sets, namely:
\begin{enumerate}
    \item \textbf{\sqhints}: follow-on hints generated from only a single related question, and
    \item \textbf{\mqhints}: follow-on hints generated from  multiple distinct related questions.
\end{enumerate}

\subsubsection{Hint Generation Guidelines}

Based on the intuitions from \S\ref{sec:linguistic_task}, we provide guidelines to annotators to create voice-friendly hints. For the tuple $\langle q, Q_{rel}\rangle$, annotators follow the steps below to write a hint.
\change{Details about the crowdsourcing setup, worker payment and hint generation quality  are provided in \S\ref{a_subsec:annotation_guidelines}}.

\noindent\textbf{Step 1.} Annotators are asked to start the hint with one of the provided \emph{start patterns} (cf. \Cref{tab:hint_requirements}).

\noindent\textbf{Step 2.a.} The questions in $Q_{rel}$ are converted into RS format. RS conversion templates are provided to annotators:
\begin{itemize}[leftmargin=*]
\itemsep-.5em
    \item \example{\textbf{Q}: Did Samuel Adams plan the Boston Tea Party?}
    \item \example{Bob wants to know if \textcolor{olive}{Samuel Adams planned the Boston Tea Party?}}
\end{itemize}

\noindent\textbf{Step 2.b.} For \mqhints, annotators need to avoid \emph{repetitions} and replace them with \emph{anaphora} where necessary. Next, subordinate clauses from $Q_{rel}$ are connected with the correct \emph{conjunctive discourse markers}, e.g. given two questions: \example{Did Samuel Adams plan the Boston Tea Party?} and \example{What was the role of Samuel Adams in the American Revolution?}, the example below shows the correct use of anaphora and conjunctions.
\begin{itemize}[leftmargin=*]
    \item \example{You may also want to know if Sam Adams planned the Boston Tea party, \textcolor{red}{or/and} about \textcolor{blue}{his} role in the American Revolution.}
\end{itemize}

\subsubsection{Data Collection}

We create two disjoint subsets: \sqhints (hints from a single related question) and \mqhints (hints for multiple questions).
Table~\ref{tab:hints_statistic} shows a detailed overview of our collected dataset. 

\begin{table}[ht!]
    \centering
    \resizebox{1.0\columnwidth}{!}{
    \begin{tabular}{l l c | l p{1.75cm} p{1.75cm}}
    \toprule
    & \multicolumn{2}{c|}{\textbf{\sqhints}} & \multicolumn{3}{c}{\textbf{\mqhints}}\\
    \midrule
    \emph{domain} & \# & ratio & \# & ratio ($|Q_{rel}| \! = \! 2$) & ratio ($|Q_{rel}| \! = \! 3$) \\
    \midrule 
\texttt{Animal} & 2,806 & - & 2,780 & 24.1\% & 75.9\% \\
\texttt{Place} & 2,105 & - & 1,369 & 3.2\% & 96.8\% \\
\texttt{Technology} & 928 & - & 897 & 5.4\% & 96.4\% \\
\texttt{Politician} & 956 & - & 766 &  8.2\% & 91.8\%\\
\texttt{Food} & 537 & - & 329 & 59.3\% & 40.7\% \\
\texttt{Athlete} & 352 & - & 209 & 16.3\% & 83.7\%\\
\texttt{Wearables} & 180 & - & 177 & - & 100\%\\
\texttt{Holiday} & 60 & - & 54 & 5.6\% & 94.4\%\\
\midrule
\textbf{total} &  7,932 & - & 6,581 & 1,132 & 5,449 \\
\bottomrule
    \end{tabular}}\vspace{-5pt}
    \caption{\small{Follow-on voice friendly hints data statistics for \sqhints and \mqhints, respectively.}}
    \label{tab:hints_statistic}
\end{table}

Our main focus is in generating hints from top--3 related questions $Q_{rel}$, however, to ensure data diversity, we also collect hints constructed from the top--1 and top--2 related questions.
This increases the utility of our dataset, as hint generation approaches must ensure hint coherence with a variable number of related questions.

As shown in Table~\ref{tab:hints_statistic}, we collect a larger sample of \sqhints. Most of it is used for pre-training of our hint generation approaches.

%% file: setup.tex
\section{Experimental Setup}
\label{sec:experiments}

We evaluate different models and assess hint quality using automatic and human evaluation metrics. Table~\ref{tab:training_data} shows the statistics about the dataset used in our experiments.

\subsection{Datasets}
\label{subsec:datasets}

\noindent\textbf{Pre-training RS Dataset.} \sqhints, which we refer to as reported speech data, is used for pretraining the hint generation approaches (cf. \S \ref{subsec:rs_pretraining}).

\noindent\textbf{Hint Generation Dataset.} For the main task of hint generation, we randomly sample questions from Table~\ref{tab:hints_statistic}, and split with 60\%/10\%/30\% for training, development, and testing. Majority of the hints are {\mqhints}, with 81\% generated from three questions, 17\% with two questions, and the remaining 2\% are {\sqhints}.  

\begin{table}[ht!]
\centering
\resizebox{0.7\columnwidth}{!}{
    \begin{tabular}{l c c c}
    \toprule
    & train & dev & test\\
    \midrule
        RS pretraining & 4,262 & 1,831 &  - \\
        Hint Generation & 4,008 & 668 & 2,005\\
        \bottomrule
    \end{tabular}}
    \caption{{Pretraining and training hint generation datasets, sampled randomly from Table~\ref{tab:hints_statistic}.}}
    \label{tab:training_data}
\end{table}

\subsection{Baselines and Approaches}
\label{subsec:approach_setup}

For all Transformer-based approaches, we experimented with \textsc{bart-base} \cite{lewis-etal-2020-bart} and \textsc{T5-base} \cite{2020t5} models. Details about model training, along with the hyperparameter setup, are provided in Appendix~\ref{sec:model_params}.

\paragraph{\textbf{Template Baseline -- \tb.}} Hints are constructed based  on manually defined templates, by first choosing a start pattern (cf. Table~\ref{tab:hint_requirements}) and then concatenating question from $Q_{rel}$ with an \emph{``or''}.

\paragraph{\textbf{Reported Speech Baseline -- \rsb.}}
We train a seq2seq model on \sqhints only, where questions are first converted into their RS format, then using \tb different questions are concatenated into a hint. \rsb represents an ablation of \ptg (only the \emph{pretraining stage}).

\noindent\textbf{Direct Hint Generation -- \dhg.} This represents our approach without pretraining. The limitation of \dhg are that it has to jointly learn all aspects of constructing voice-friendly hints, which may lead to cases where subordinate clauses are not in the desired syntax, or the hint lacks coherence. %

\paragraph{\textbf{Hint Generation with RS Pretraining -- \ptg.}} This represents our final approach with pretraining on the RS task. Breaking down the training into two stages, \ptg first learns RS rewriting, then it learns to avoid \emph{repetitions} and ensure hint coherence and right order of subordinate clauses.

\subsection{Evaluation Metrics}
\label{subsec:evaluation_metrics}

Evaluating hint quality is not trivial. Given the task novelty and the lack of metrics that capture voice-friendliness, we opt for a combination of automatic metrics and human evaluations.

\subsubsection{Automated Metrics} 
\label{subsec:automatic_eval}

To assess the closeness of the generated hints with respect to their ground-truth counterparts generated by human annotators, we use BLEU \cite{papineni-etal-2002-bleu}, ROUGE \cite{lin-2004-rouge} and F1-BertScore \cite{Zhang*2020BERTScore:}. BLEU captures the accuracy in terms of the \emph{n-grams}, whereas ROUGE quantifies coverage of the ground-truth \emph{n-grams} in the generated hint. Finally, BERTScore computes the semantic similarity between two hints, thus accounting for the use of equivalent phrases or synonyms in the hints.

\subsubsection{Human Evaluation}
\label{subsec:eval_scenarios}

Automated metrics are good quality indicators, but they do not capture hint voice-friendliness. We devise a set of human evaluations which judge the correctness and naturalness of a hint.
For a realistic evaluation, all human studies\footnote{\emph{Question coverage}  and \emph{syntactic correctness} are done in text, as the annotators need to map questions to hints.} are performed in a voice modality.\footnote{We use Amazon's AWS Polly text-to-speech service to convert the generated texts into spoken utterances.}
We consider the following studies: (i) syntactic correctness, (ii) input question coverage,  (iii) hint pairwise comparison from different approaches, and (iv) question retention.

\noindent\textbf{Syntactic Correctness.} Annotators judge whether a hint is \emph{syntactically correct}, and if the hint uses \emph{idiomatic} expressions in English.

\noindent\textbf{Question Coverage.} Given a hint $h$ and $Q_{rel}$, annotators assess if $h$ covers all questions in $Q_{rel}$.

\noindent\textbf{Pairwise Hint Comparison.}
For two generated hints $h_a$ and $h_b$ from the same set of questions $Q_{rel}$ and two different approaches, annotators choose their preferred hint. To reduce any positional bias, hints are ordered randomly. %
\change{Finally, for each comparison we collect three judgements, achieving an inter-annotator absolute agreement rate of 0.77.}

\noindent\textbf{Question Retention.} 
We consider retention of a hint's information in memory as a proxy for its simplicity and comprehensibility. Hints cannot be considered actionable if listeners cannot remember them. 
We assess how well annotators can recall the conveyed information in a hint and ask questions about \emph{one} of the conveyed topics in the hint.
To emulate interaction with a voice assistant, annotators first listen to the hint, after which a mandatory 5 seconds pause is enforced. Then they need to choose the correct question covered in $h$ from a set of four questions shown to them. Only one of the questions is present in $h$. We select the three distractor questions, one chosen at random, and the other two are either relevant to the entity and topic covered by $h$, or the entity only.

%% file: evaluation.tex
\section{Evaluation on Automated Metrics}
\label{sec:evaluation}

\begin{table*}[ht!]
    \centering
    \resizebox{.9\textwidth}{!}{\small{
    \begin{tabular}{l l l l l l l l l l}
    \toprule
         & {\small BLEU1} & {\small BLEU2}& {\small BLEU3} & {\small BLEU4} & {\small ROUGE1} & {\small ROUGE2}& {\small ROUGE3} & {\small ROUGE4} & {\small BERTScore}\\ 
        \midrule
        \tb & 0.509 &	0.401 & 	0.323 &	0.254 &	0.713 &	0.488 &	0.358 &	0.278 &	0.536 \\
        \rsb & 0.519 & 0.415 &	0.341 &	0.274 &	0.717 &	0.501 &	0.375 &	0.292 &	0.494\\
        \dhg-T5 & 0.616 &	0.510 &	0.428 &	0.358 &	0.728 &	0.525 &	0.400 &	0.320 &	0.632 \\
        \dhg-BART & 0.616 &	0.509 &	0.427 &	0.359 &	0.734 &	0.529 &	0.402 &	0.322 &	0.628\\
        \ptg-T5 & 0.629 &	0.524 &	0.442 &	0.373 &	0.739 &	0.534& 	0.410 &	0.329 &	\textbf{0.643}\\
        \ptg-BART & \textbf{0.630} &	\textbf{0.527} &	\textbf{0.446} &	\textbf{0.378} &	\textbf{0.742} &	\textbf{0.539} &	\textbf{0.413} &	\textbf{0.333} &	0.642\\
        
        \bottomrule
    \end{tabular}}}
    \caption{
    {\ptg-BART achieves the highest performance across nearly all evaluation metrics, obtaining statistically highly significant results ($p<0.01$) against all its counterparts. \Cref{sec:llm_hints} shows a comparison between \ptg-BART and ChatGPT.}}
    \label{tab:automatic}
\end{table*}

\begin{table*}[ht!]
    \centering
    \resizebox{1.0\textwidth}{!}{
    \begin{tabular}{l l l l l l l l l l l l l l l l l l l}
    \toprule
        \textbf{Domain} & \multicolumn{2}{c}{BLEU1} &  \multicolumn{2}{c}{BLEU2}&  \multicolumn{2}{c}{BLEU3} &  \multicolumn{2}{c}{BLEU4} &  \multicolumn{2}{c}{ROUGE1} &  \multicolumn{2}{c}{ROUGE2}&  \multicolumn{2}{c}{ROUGE3} &  \multicolumn{2}{c}{ROUGE4} &  \multicolumn{2}{c}{BertScore}\\ 
        \midrule
        & \dhg & \ptg & \dhg & \ptg & \dhg & \ptg & \dhg & \ptg & \dhg & \ptg & \dhg & \ptg & \dhg & \ptg & \dhg & \ptg & \dhg & \ptg \\
        \midrule
        
\texttt{Animal} &  0.604 & \textbf{0.618}$^{\ddag}$ & 0.498 & \textbf{0.511}$^{\ddag}$ & 0.413 & \textbf{0.426}$^{\ddag}$ & 0.344 & \textbf{0.356}$^{\ddag}$ & 0.717 & \textbf{0.732}$^{\ddag}$ & 0.509 & \textbf{0.521}$^{\ddag}$ & 0.377 & \textbf{0.388}$^{\ddag}$ & 0.298 & \textbf{0.308}$^{\ddag}$ & 0.634 & \textbf{0.642}$^{\ddag}$\\

\texttt{Athlete} & 0.563 & \textbf{0.573}$^{\dagger}$ & 0.440 & \textbf{0.450} & 0.345 & \textbf{0.356} & 0.261 & \textbf{0.275} & 0.696 & \textbf{0.705} & 0.478 & \textbf{0.488} & 0.347 & \textbf{0.357} & 0.266 & \textbf{0.276} & \textbf{0.587} & 0.586\\

\texttt{Food} & 0.636 & \textbf{0.643}$^{\ddag}$ & 0.540 & \textbf{0.550}$^{\ddag}$ & 0.466 & \textbf{0.478}$^{\ddag}$ & 0.401 & \textbf{0.413}$^{\ddag}$ & 0.749 & \textbf{0.752} & 0.561 & \textbf{0.568}$^{\dagger}$ & 0.443 & \textbf{0.453}$^{\ddag}$ & 0.364 & \textbf{0.374}$^{\dagger}$ & 0.686 & \textbf{0.695}$^{\ddag}$\\

\texttt{Holiday} &  0.572 & \textbf{0.586} & 0.459 & \textbf{0.472} & 0.369 & \textbf{0.387} & 0.289 & \textbf{0.320}$^{\dagger}$ & 0.717 & \textbf{0.727} & \textbf{0.511} & 0.504 & \textbf{0.399} & 0.395 & \textbf{0.327} & 0.325 & 0.592 & \textbf{0.608}$^{\dagger}$\\

\texttt{Places} &  0.602 & \textbf{0.619}$^{\ddag}$ & 0.493 & \textbf{0.512}$^{\ddag}$ & 0.411 & \textbf{0.431}$^{\ddag}$ & 0.341 & \textbf{0.360$^{\ddag}$} & 0.754 & \textbf{0.760}$^{\dagger}$ & 0.550 & \textbf{0.554} & 0.427 & \textbf{0.429} & 0.342 & \textbf{0.343} & 0.599 & \textbf{0.617}$^{\ddag}$\\

\texttt{Politician} &  0.561 & \textbf{0.579}$^{\ddag}$ & 0.440 & \textbf{0.459}$^{\ddag}$ & 0.348 & \textbf{0.370}$^{\ddag}$ & 0.277 & \textbf{0.298}$^{\ddag}$ & 0.715 & \textbf{0.719}$^{\dagger}$ & 0.491 & \textbf{0.494} & 0.360 & \textbf{0.362} & 0.283 & \textbf{0.284} & 0.552 & \textbf{0.573}$^{\ddag}$\\

\texttt{Technology} &  0.598 & \textbf{0.608}$^{\ddag}$ & 0.484 & \textbf{0.497}$^{\ddag}$ & 0.398 & \textbf{0.411}$^{\ddag}$ & 0.329 & \textbf{0.340}$^{\ddag}$ & 0.738 & \textbf{0.750}$^{\ddag}$ & 0.537 & \textbf{0.550}$^{\ddag}$ & 0.418 & \textbf{0.427}$^{\ddag}$ & 0.342 & \textbf{0.348}${\dagger}$ & 0.576 & \textbf{0.589}$^{\ddag}$\\

\texttt{Wearables} &  0.658 & \textbf{0.659} & 0.572 & \textbf{0.573} & 0.505 & \textbf{0.506} & 0.447 & \textbf{0.448} & 0.794 & \textbf{0.795} & \textbf{0.620} & 0.619 & \textbf{0.502} & 0.499 & \textbf{0.425} & 0.424 & 0.626 & \textbf{0.628}\\
        \bottomrule
    \end{tabular}}
    \caption{
    {Comparison on out-of-domain hint generation performance for  \ptg-BART and \dhg-BART. With $^{\dagger}$ are denoted statistically significant ($p <0.05$) and with $^{\ddag}$ highly significant results  ($p < 0.01$). }}
    \label{tab:automatic_domains}
\end{table*}

Table~\ref{tab:automatic} shows the performance measured on the automated metrics for the different approaches.

\paragraph{Baseline Performance:} \tb achieves the lowest scores across all metrics (except for BERTScore). This is expected, since concatenated questions are compared w.r.t the ground-truth hints, written by annotators. \rsb obtains a consistent improvement across all metrics. It rewrites individual questions into content clauses, which then are concatenated using the conjunction \example{or}. However, \rsb does not reduce lexical repetition via anaphora, and simple concatenation results in lower coherence.
Overall \tb and \rsb, achieve low scores as expected. More insights are provided by the human evaluation studies, which capture hint voice friendliness.

\paragraph{Approach Performance:}
Our approaches, \dhg and \ptg, show a consistent improvement over \tb and \rsb across all automated metrics. This is intuitive since they are optimized to generate hints.

Comparing \ptg and \dhg in Table~\ref{tab:automatic}, we note a \emph{significant} improvement in terms of BLEU scores due to the pretraining phase. This follows our intuition that pretraining helps \ptg to convert questions into subordinate clauses, a key aspect of natural hints. In the fine-tuning stage, \ptg can already reasonably convert questions into RS syntax, and thus can focus on reducing lexical redundancy, resulting in more coherent hints. While \ptg employs multi-stage training, in \dhg all operations are learned end-to-end. This represents a complex training regime, requiring optimization of several rewrite tasks, listed in Table~\ref{tab:hint_requirements}. 

The difference in performance between \ptg and \dhg, demonstrates that for complex  rewriting tasks, end-to-end training may be sub-optimal. Decomposing the problem into specific pretraining subtasks before fine-tuning in an end-to-end manner yields significant improvements. Similar finding are reported in~\cite{DBLP:conf/interspeech/AroraO0DM0B21}. 

For ROUGE metrics, only \ptg-T5 obtains significantly better results than \dhg-T5 for ROUGE1. For the rest, although \ptg has higher ROUGE scores, the differences are not significant. Finally, for BERTScore the differences are significant between \ptg-BART over \dhg-BART.

\paragraph{Robustness:}
Table~\ref{tab:automatic_domains} shows an out-of-domain evaluation, for \ptg-BART and \dhg-BART. This assesses model robustness on unseen domains during training. 
Comparing the performance of \ptg-BART and \dhg-BART, we note that across all domains, pretraining in \ptg allows the model to achieve significantly better results than \dhg. Only for  \texttt{Wearables} do we not observe any significant difference. This can be attributed to the smaller test set size, with only 45 instances. Additional evaluation results are shown in Appendix~\ref{a_subsec:robustness}.

\section{Human Evaluation Studies}

\subsection{\change{Syntactic Correctness} and Coverage}

Table~\ref{tab:properties} shows the performance of the different models in terms of input questions coverage and the syntactic correctness. For a random sample of 500 hints and the corresponding $Q_{rel}$, we assess if all input questions are present in a generated hint, and if the hint is \change{syntactically correct}. 
\begin{table}[ht!]
\centering
\resizebox{1\columnwidth}{!}{
\begin{tabular}{l c c}
\toprule
    \textbf{Approach} & \textbf{Syntactic Correctness} & \textbf{Question Coverage} \\ 
    \midrule
    \tb & 449 (89.8\%) & 500  (100\%)\\
    \rsb & 461 (92.2\%) & 484 (96.8\%)\\
    \dhg-T5 & 434 (86.8\%) & 464  (92.8\%)\\
    \dhg-BART & 428 (85.6\%)& 466 (93.2\%)\\
    
    \ptg-T5 & 431 (86.2\%) & 485 (97.0\%)$^{\ddag}$ \\
    \ptg-BART & 455 (91.0\%)$^{\ddag}$ & 485 (97.0\%)$^{\ddag}$\\
    \bottomrule
\end{tabular}}
\caption{
{Syntactic correctness and input question coverage results. Significant differences between \ptg and \dhg  are marked with $\ddag$. No significant difference exists between \dhg and \rsb (as per binomial test of proportions).} }
\label{tab:properties}
\end{table}

\textbf{Syntactic Correctness.}
Table~\ref{tab:properties} shows a consistent pattern in terms of syntactic correctness: the baseline \rsb and \ptg-BART have the highest portion of syntactically correct hints as judged by the annotators, with 92\% and 91\%, respectively. Generating hints from multiple questions is not trivial, as it involves syntactic and stylistic changes in $h$, allowing room for errors for generative models, especially in terms of syntactic errors.

The high \rsb and \ptg-BART scores can be interpreted as follows. \rsb is trained on \sqhints, which does a syntactic conversion of the input question into their RS format, and through simple rules concatenates content clauses. This allows the model to generate hints that are syntactically correct in 92\% of the cases. Similarly, \ptg-BART, that is pretrained on \sqhints, has the same capabilities as \rsb, and generates in 91\% of the cases syntactically correct hints. However, contrary to \rsb, \ptg-BART additionally fine-tunes for voice-friendliness, which ensure hint coherence and redundancy. While \rsb generates syntactic hints, its hints are far less natural than those of \ptg-BART (cf. \S \ref{subsubsec:pairwise}).
    
\textbf{Coverage.} For question coverage, we note that the \ptg approaches achieve the highest coverage among the learning based approaches, with 97\% of the hints covering all the questions. \tb has perfect coverage, given that its hints are generated by simply concatenating the input questions. 

Finally, the \dhg approaches have the lowest coverage, with 92.8\% of hints having full coverage. This indicates that end-to-end learning of all hint generation tasks is challenging.

\subsection{Pairwise Hint Comparison}
\label{subsubsec:pairwise}

Here we measure which approaches generate hints that are considered more natural by humans. As \dhg has consistently lower performance than \ptg, we only compare \ptg-BART, \rsb, and \tb. 
To understand the naturalness of the hints in a spoken format, they are converted to audio. After listening to the hints, annotators judge which hint they find more \emph{natural} and \emph{easier to understand}. To avoid positional bias, the order in which the hints are played is randomized. 

Table~\ref{tab:pairwise} shows the pairwise comparisons the different models. We run the comparison on the 441 hints that were judged to be syntactically correct in \Cref{tab:properties}. This is done to avoid any bias stemming from syntactically incorrect hints.
\begin{table}[ht!]
\centering
\resizebox{1\columnwidth}{!}{
\begin{tabular}{l c c}
    \toprule
    \textbf{Comparison} &  \textbf{\ptg-BART chosen} & \textbf{Baseline chosen}\\ 
    \midrule
    \ptg-BART vs. \tb & 300 (68\%) & 141 (32\%)\\
    \ptg-BART vs. \rsb & 267 (61\%) & 174 (39\%)\\
    \bottomrule
\end{tabular}}
\caption{{Pairwise hint comparison. \ptg-BART hints are significantly ($p<0.01$, as per binomial test of proportions) considered to be more voice-friendly than the baselines hints. }}
\label{tab:pairwise}
\end{table}

In both comparisons, \ptg-BART produces more natural hints than baselines. Against \tb, it is preferred in 68\% of the cases, whereas against \rsb, this is in 60\% of the cases. Both results represent statistically highly significant differences (as per Wilcoxon's signed-rank test). %
\Cref{tab:pairwise_domain} shows the pairwise comparison at the domain level, for all domains \ptg-BART is preferred by human annotators as having more voice friendly hints.

\begin{table}[ht!]
\centering
\resizebox{.9\columnwidth}{!}{
\begin{tabular}{ l r r}
\toprule
    \textbf{Domain} & \textbf{\ptg-BART vs. \tb} &  \textbf{\ptg-BART vs. \rsb}\\
    \midrule
    \texttt{Animal} & 135/69 & 122/82\\
    
    \texttt{Places} & 59/20 & 43/36\\
    \texttt{Tech} & 39/20 & 37/22\\
    \texttt{Politician} & 30/15 & 30/15\\
    \texttt{Food} & 14/10 & 15/9\\
    
    \texttt{Athlete} & 10/6 & 11/5\\
    \texttt{Wearables} & 10/1 & 7/4\\
    \texttt{Holiday} & 3/0 & 2/1\\ 
    
    \bottomrule
\end{tabular}}
\caption{{Per-domain pairwise hint comparison results.}}
\label{tab:pairwise_domain}
\end{table}

\subsection{Question Retention Evaluation}
\label{subsubsec:hint_retention}

In the final human evaluation from \S \ref{subsec:eval_scenarios}, we measure how actionable the generated hints are. Beyond being natural or correct, the main aim of generating follow-on hints is for them to be actionable such that listeners (\ie users of voice assistants) can ask follow-up questions. 

Using the same set of 441 syntactically correct hints (cf. Table~\ref{tab:properties}), annotators listen to the hints, after which a set of four questions is shown, where only one was actually part of the hint. The ability to correctly \textit{recognize} this question is a proxy for whether the listeners could comprehend and remember the hint's information content.\footnote{More details about hint length/retention are in \S\ref{a_subsec:robustness}} In a conversational scenario with a voice assistant, they could follow-up by asking this question.

Table~\ref{tab:retention} shows the retention for different approaches. \ptg-BART and \dhg-BART achieve significantly better retention than the baselines \tb and \rsb.
This finding demonstrates that retention is negatively impacted by incoherent (\tb due to simple concatenation) and repetitive (\rsb due to it not using anaphora) hints.

\begin{table}[ht!]
\centering
\resizebox{1\columnwidth}{!}{
\small{
\begin{tabular}{l p{2cm} p{2cm}}
\toprule
    \textbf{Model} & \textbf{\makecell{\# Recognized\\ Questions}}  & \textbf{\makecell{Hint Length\\ (\# characters)}}\\ 
    \midrule
    Templates (TB) & 356 (80.7\%) & 152.72 $\pm$ 34.6\\
    \rsb & 348 (78.9\%) & 158.02 $\pm$ 34.6\\
    \dhg-BART & 383 (86.8\%) & 139.85 $\pm$ 33.8\\
    \ptg-BART & 384 (87.1\%) & 140.78 $\pm$ 34.5\\
    \bottomrule
\end{tabular}}}
\caption{
{Number of hints correctly recognized as being part of the hint by annotators, who selected between four questions, where only one is correct.}}
\label{tab:retention}
\end{table}

%% file: related_work.tex
\section{Related Work}\label{sec:related_work}

Our task is novel and thus has no directly comparable works, closest being on question generation.

\paragraph{\textbf{Question Generation.}} Rus et al.~\shortcite{rus-etal-2010-first} for a given input paragraph generate questions. The works in  \cite{10.1007/978-3-319-13704-9_5,8609569} make use of knowledge graphs (KG) and predefined templates, such as \example{what is X}, where X is some entity from the KG.  

Rosset et al.~\shortcite{10.1145/3366423.3380193} propose an approach for \emph{conversational} question generation based on the GPT-2~\cite{radford2019language}. Given a user question, a follow-on question is suggested to the user, that can be seen as a continuation of their search trajectory. Rao et al.~\shortcite{b-etal-2020-automatic} generate follow-up questions for interviews, where after a question, an answer, a follow-up question is generated.  

Our approach can be seen related to these works, especially to \cite{10.1145/3366423.3380193} given that we both aim at increasing user engagement. Yet, we differ in two fundamental ways: 1) we do generate questions but hints about questions that can be asked, and 2) through hints we allow users to explore additional topics. Finally, we do not focus on \emph{what} but rather \emph{how} to generate hints.

\paragraph{\textbf{Conversational Text Generation.}} Su et al.~\shortcite{DBLP:conf/acl/SuSZZHZNZ20} propose a pretraining approach for diversifying seq2seq models in generating non-conversational text for dialogues, by additionally training on non-conversational text extracted from books. Similarly, in \cite{DBLP:conf/acl/ZhangSGCBGGLD20} a GPT-2 model is pretrained over Reddit conversation chains. Targeted conversational question generation approaches~\cite{DBLP:conf/acl/PanLYCS19,DBLP:conf/eacl/GuMYS21} take into account the conversation history and the topic of interest, and  generate possible next questions that can be answered. These methods deal with how to generate conversational text, and thus are very different to our use case. Past works on follow-up conversation turn generation, either considers a question~\cite{DBLP:conf/acl/PanLYCS19,DBLP:conf/eacl/GuMYS21} or other non-conversational snippet~\cite{DBLP:conf/acl/ZhangSGCBGGLD20}, and focus on generating snippets that are extracted from a single sentence or passage, thus not directly dealing with text coherence. Additionally, no voice-friendly aspects are considered,  diminishing their utility on voice assistants.

\paragraph{\textbf{Text Summarization.}} Generating compact summaries from lengthy documents has been the focus of various approaches~\cite{kryscinski-etal-2019-neural}. Abstract text summarization ~\cite{DBLP:conf/emnlp/JiangB18,DBLP:conf/iclr/PaulusXS18,durrett-etal-2016-learning} are typically deployed in scenarios where the input text needs to be \emph{summarized} and at the same time \emph{paraphrased}. On the contrary, our task, instead of paraphrasing, requires \emph{stylistic} changes such as rewriting questions in their \emph{indirect speech} form. Moreover, instead of summarizing, our task entails \emph{syntactic} changes, such as use of pronouns to avoid redundancy, and coordinating the different subordinate clauses using conjunctive phrases. The two tasks have inherently different aims and as such require optimizing for different objectives.
\change{We experimented with several pre-trained summarization models, however, expectedly their performance was poor, thus, do not include those results as baselines in the paper.}

\paragraph{\textbf{Paraphrasing.}} \change{Related works on paraphrasing~\cite{witteveen-andrews-2019-paraphrasing,niu-etal-2021-unsupervised,bannard-callison-burch-2005-paraphrasing} make use of pre-trained language model to paraphrase input sentences into semantically equivalent sentences, which make use of different phrases and wording. The main difference of our task to paraphrasing lies in combining different interrogative clauses from related questions into a coherent hint, while paraphrasing does not enforce strict syntactic patterns as required in voice friendly hints (cf. Table~\ref{tab:hint_requirements}). }

\paragraph{\textbf{Evaluation Metrics.}} Guy~\shortcite{10.1145/3182163} in his analysis of spoken and Web search queries identifies that voice questions have phonetic properties such as \emph{speed} and \emph{intonation} that are not present in text queries. This poses challenges when using automated metrics such as BLEU, ROGUE, where the output of a model is \emph{voice}, but it is trained on text data. Similar to the work in \cite{mehri-eskenazi-2020-unsupervised}, which introduces several task specific evaluation metrics to measure dialog quality, e.g. \emph{fluency}, \emph{engagement}, \emph{correctness}, we follow a similar strategy and propose several human evaluations to measure voice friendliness of a hint.

%% file: conclusions.tex
\section{Conclusions}
\label{sec:conclusions}

We presented a novel approach for question suggestion using spoken hints.
Our work enables the creation of new voice-based experiences where users can receive compact and natural hints about additional questions they can ask.
Question suggestion is a standard feature in screen-based search experiences, and our work takes an important first step in bringing this capability to voice interfaces.

Our contributions are manifold: (i) a novel task of suggesting questions with voice hints; (ii)  outlined the linguistic desiderata and processes to decompose questions into interrogative content clauses, and recompose  them into declarative hints; and (iii) a new dataset of over $14$k input questions and hints, using carefully constructed annotation guidelines and quality checks.

We defined seq2seq models to generate hints.
Using both automatic metrics and human evaluations, we conclusively showed that our most sophisticated approach \ptg, which utilizes a linguistically motivated pretraining task was strongly preferred by humans with most natural hints.

%% file: appendix_v2.tex
\section*{\centering Appendix}

The appendix contains details about the question data collection, crowdsourcing job setup for the hint data annotation task, and more detailed model evaluation on different aspects, such as examples and approach robustness. 

Finally, it contains the workflows of our voice-friendly hint generation approaches.

\section{Question Suggestion via Voice Friendly Hints}\label{a_subsec:linguistic_task}

\begin{figure*}[ht!]
    \centering
    \includegraphics[width=1\textwidth]{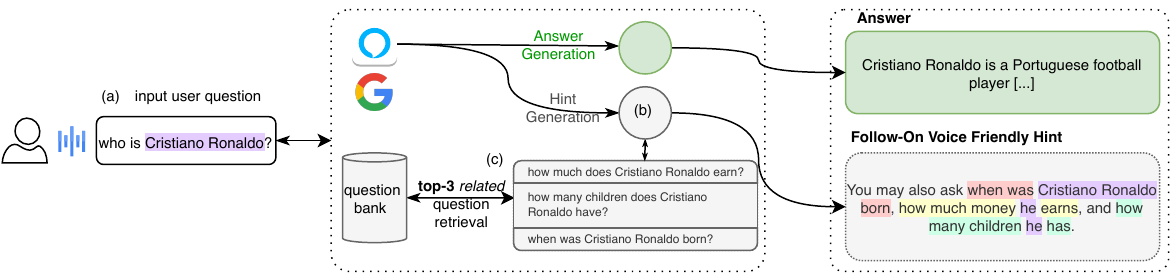}\vspace{-10pt}
    \caption{\small{Voice-friendly Hint Generation Task: (a) for an input user question, the voice-assistant generates a voice-friendly hint (c) from top-3 related questions about the entity in (a) retrieved from its question bank.}}
    \label{fig:task}
\end{figure*}

In Figure~\ref{fig:task} are shown the different steps that are invoked when a user interacts with a voice assistant in order to obtain related question suggestions via voice friendly hints.

\section{Question \& Related Question Retrieval}\label{a_subsec:question_dataset}

\textbf{Input Questions:} We first create a pool of input questions by extracting $521$k questions from the MS-MARCO dataset~\citep{msmarco}, extracted from sentences that start with \emph{wh--*} phrases from community QA websites (e.g. Quora, Yahoo Answers). By limiting to QA community pages, we can get a diverse and high quality questions.

\subsection{Related Question Retrieval}\label{a_subsec:preprocessing}
For $q$ to retrieve its related questions $Q_{rel}$, we first determine the entity and topic of a question. 

\textbf{Entity Linking:} Entities are extracted from questions using the Blink approach \cite{blink}, which are then mapped to their \emph{types} from DBpedia~\cite{dbpedia}, restricted to only the following domains: $D=\{$\domain{Animal}, \domain{Athlete}, \domain{Food}, \domain{Holiday}, \domain{Places}, \domain{Politician}, \domain{Technology}, \domain{Wearables}, \domain{Video Game}$\}$.

\textbf{Topic Extraction:} As topics we consider the \emph{predicates} associated to entities in DBpedia. For instance, $T_d=$\{\emph{``birth place''} $\ldots$ \emph{``death place''}\} are extracted from entities of domain \domain{Politician}. A question $q$ is associated to a predicate from $T_d$ based on the highest semantic similarity~\cite{GoogleUSE} between the topic and question keywords (and enforce a minimum threshold cosine similarity of 0.1). %
Finally, the question bank becomes the set of quadruples $Q=\{\langle q, e, d, t\rangle_1, \ldots \langle q, e, d, t\rangle_n\}$.%

\paragraph{Related Question Retrieval:} For an input question $q_i\in Q$, by filtering the quadruples in $Q$ we obtain the top--$k$ related questions $Q_{rel}$ as the questions that have the same subject entity as $q_i$ and which cover a different topic from $q_i$, namely, $Q_{rel} = \{ \langle q, e, d, t\rangle_j | e_j = e_i \wedge t_j \neq t_i\}, \forall j \leq |Q|$. The top--$k$ ranges with $k=\{1,2,3\}$, chosen from most frequent topics in $Q$.

\section{Hint Annotation Guidelines}\label{a_subsec:annotation_guidelines}

\begin{figure}
    \centering
    \frame{\includegraphics[width=1.0\columnwidth]{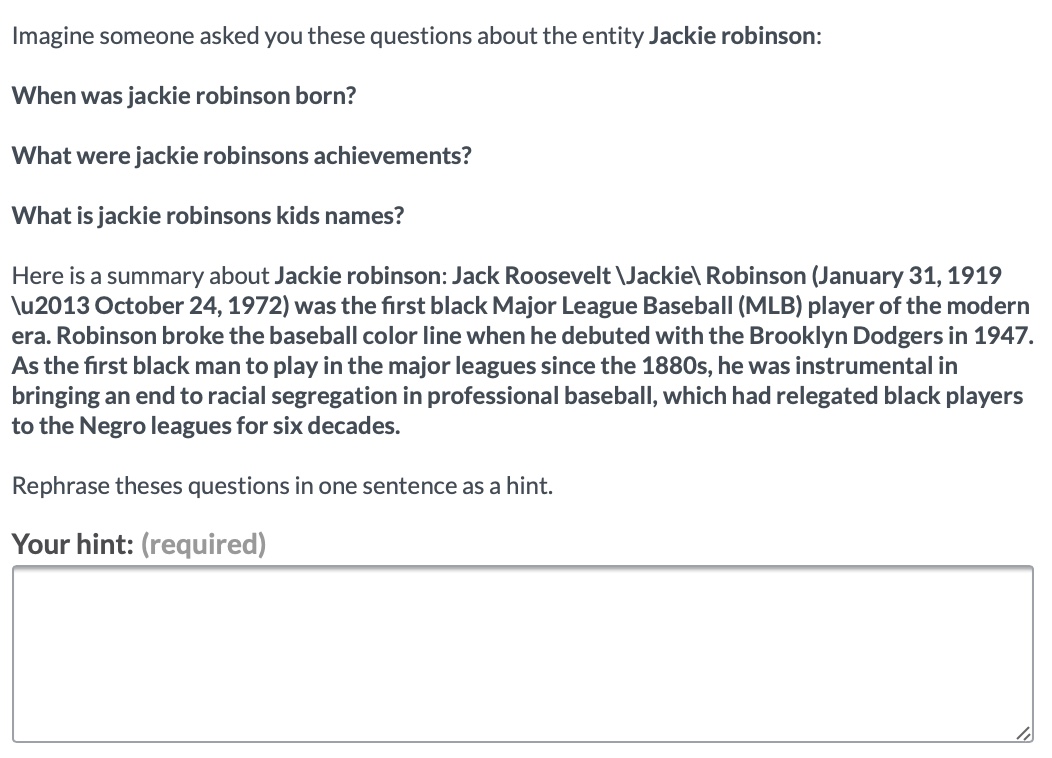}}
    \caption{\small{Annotation interface for obtaining voice-friendly hints showing three related questions about the entity \emph{``Jackie Robinson''} along with a short summary of the entity itself, extracted from Wikipedia.}}
    \label{fig:annotation_view}
\end{figure}

Annotators from the Appen crowdsourcing platform\footnote{\url{https://appen.com}} are given the related questions $Q_{rel}$, and asked to compose a corresponding hint $h$. \change{Figure~\ref{fig:annotation_view} shows the annotation interface, while the guidelines and steps are explained in the following.} %

We rely only on annotators  with highest level of competence\footnote{Level 3 workers in Appen} that were also English native speakers, and paid  according to the time spent in a task at a rate of \$15 (USD) per hour.\footnote{This is the minimum hourly wage in WA, USA}

\change{Finally, we enforce a set of validation mechanisms to avoid malicious behavior from annotators. Table~\ref{tab:validators} shows the set of validators used to ensure quality of obtained annotations.} 
\change{Any generated hint that does not meet any of the validators in the table below is discarded. Furthermore, hints are run through Gramformer\footnote{\tiny\url{https://github.com/PrithivirajDamodaran/Gramformer}} to correct any potential grammar mistakes by the human annotators.}

\begin{table}[h!]
    \centering
    \resizebox{1.0\columnwidth}{!}{
    \begin{tabular}{p{4cm} p{5cm}}
    \toprule
    Minimum/maximum \emph{characters} per hint\tablefootnote{Minimum of 70 characters, and a maximum that does not exceed the number of characters from the input question.}     &  Minimum/maximum \emph{words} per hint\tablefootnote{We set the minimum and maximum number of words to be related to the length of input questions.} \\
    \emph{Hint Coherence}\tablefootnote{Use Appen's natural language coherence functionality and use a minimum threshold, which the annotators need to pass in order for them to proceed further in the task.} & \emph{Language} constraint (\texttt{EN})\\
    Presence of \emph{start pattern}     &  Presence of \emph{entity}\\
    Presence of \emph{anaphora} & Hint/Question(s) similarity\tablefootnote{Cosine similarity computed between the generated hint and the input questions.} \\
    \bottomrule
    \end{tabular}}
    \caption{\small{Validation mechanisms to ensure data quality.}}
    \label{tab:validators}
\end{table}

\section{Model Setup \& Hyperparameters}\label{sec:model_params}

For all transformer based models, namely, \tb, \dhg, and \ptg, we use the following hyper-parameters for model training. We consider a learning rate of $lr=3e^{-5}$ with a weight decay of $d=0.01$. We train the model with a maximum 50 epochs and batch size of 8. The training stops after 10 epoch of non-decreasing validation loss.

\section{Hint Generation Performance}\label{a_subsec:hint_gen_performance}

The examples below show hints for the same set of questions, generated from all competing approaches. Depending on the questions in $Q_{rel}$, \tb may or may not produce voice-friendly hints, as the questions are simply concatenated using templates. For \rsb on the other hand we see that it does a series of rewrites. The differences between \dhg and \ptg are subtle, such as, rewriting \example{does earn} $\rightarrow$ \example{earns}, which is attributed to \ptg's pretrained RS knowledge. This allows the model to express the same information with fewer words and in a more voice-friendly manner.

\begin{table}[h!]
\centering
\resizebox{1\columnwidth}{!}{
    \footnotesize{
    \begin{tabular}{|p{0.6cm}|p{8.5cm}|}
    \hline
    \rowcolor{gray!10}
    \textbf{\footnotesize{$Q_{rel}$}} & \emph{How much money does Cristiano Ronaldo earn? How many children does Cristiano Ronaldo have? Who is the mother of Cristiano Ronaldos child?}\\\hline\hline
    
    \rowcolor{red!10}
    \textbf{\footnotesize{\tb}} & \emph{You may want to know how much money does Cristiano Ronaldo earn, or how many children does Cristiano Ronaldo have, or who is the mother of Cristiano Ronaldos child.}\\\hline
    
    \rowcolor{orange!10}
    \textbf{\footnotesize{\rsb}} & \emph{You may want to know how much money Cristiano Ronaldo earns, or how many children Cristiano Ronaldo has, or who is the mother of Cristiano Ronaldo child.}\\\hline
    
    \rowcolor{yellow!25}
    \textbf{\footnotesize{\dhg}} &  \emph{You may want to know how much money does Cristiano Ronaldo earn, or how many children he has, or who is the mother of his child.} \\\hline
    
    \rowcolor{green!10}
    \textbf{\footnotesize{\ptg}} &  \emph{You may want to know how much money Cristiano Ronaldo earns, or how many children he has, or who is the mother of his child.	}\\

    \hline
    \end{tabular}}}
    
\end{table}

\subsection{Approach Robustness}\label{a_subsec:robustness}

\begin{figure}
    \centering
    \includegraphics[width=1.0\columnwidth]{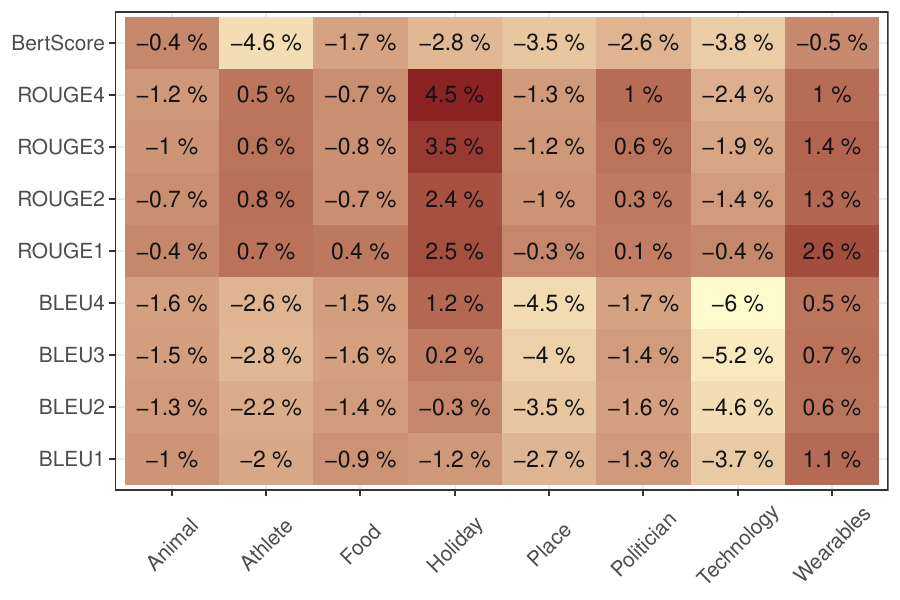}\vspace{-10pt}
    \caption{\small{Performance gap of \ptg-BART when evaluated in a zero-shot setting on a target domain (not seen during training) when compared to its performance when the model has been trained on questions from the target domain.}}
    \label{fig:performance_gap}
\end{figure}

Figure~\ref{fig:performance_gap} shows the gap in terms of performance across the different evaluation metrics for the \ptg-BART when applied in a zero-shot setting on a \emph{target} domain, compared to when the model is trained with questions from that domain. We note that overall, the gap is quite small, with many domains having a gap of 1-2\%, with th exception of \texttt{Holiday, Place} and \texttt{Technology}, which have higher gaps.
Such results show a promising generalization of \ptg-BART across domains, an indicator that the models effectively learn how to perform the various syntactic operations (cf. Table~\ref{tab:hint_requirements}) to produce voice-friendly hints. 

\paragraph{Hint Length vs. Question-Retention:} We measure the Pearson correlation between hint length (in characters) and the question retention from the generated hints. We note a negative moderate correlation of $\rho=-0.47$ between length and retention rate. Longer hints impact annotators' comprehension performance, resulting in their inability to correctly identify the suggested question in the hint.  
This confirms our hypothesis, that a key aspect to voice-friendliness such as length, has  a negative impact in a conversational setting between user and a voice-assistant in consuming such hints.

\subsection{Hint Examples}

Table~\ref{tab:sentences_examples} shows hints generated from the different competing approaches on the same set of input questions.

\section{Large Language Models for Voice Friendly Hint Generation}\label{sec:llm_hints}

Large language models (LLMs) like GPT3.5 or ChatGPT,\footnote{\url{https://chat.openai.com}} which leverage billions of parameters, are shown to have great zero-shot capabilities for various tasks in NLP. While, LLMs are impractical in our setting, where the latency requirements make it nearly impossible to use such models, nonetheless we compared our models \ptg and \dhg against ChatGPT. 

\begin{figure}[h!]
\centering
\includegraphics[width=1.0\columnwidth]{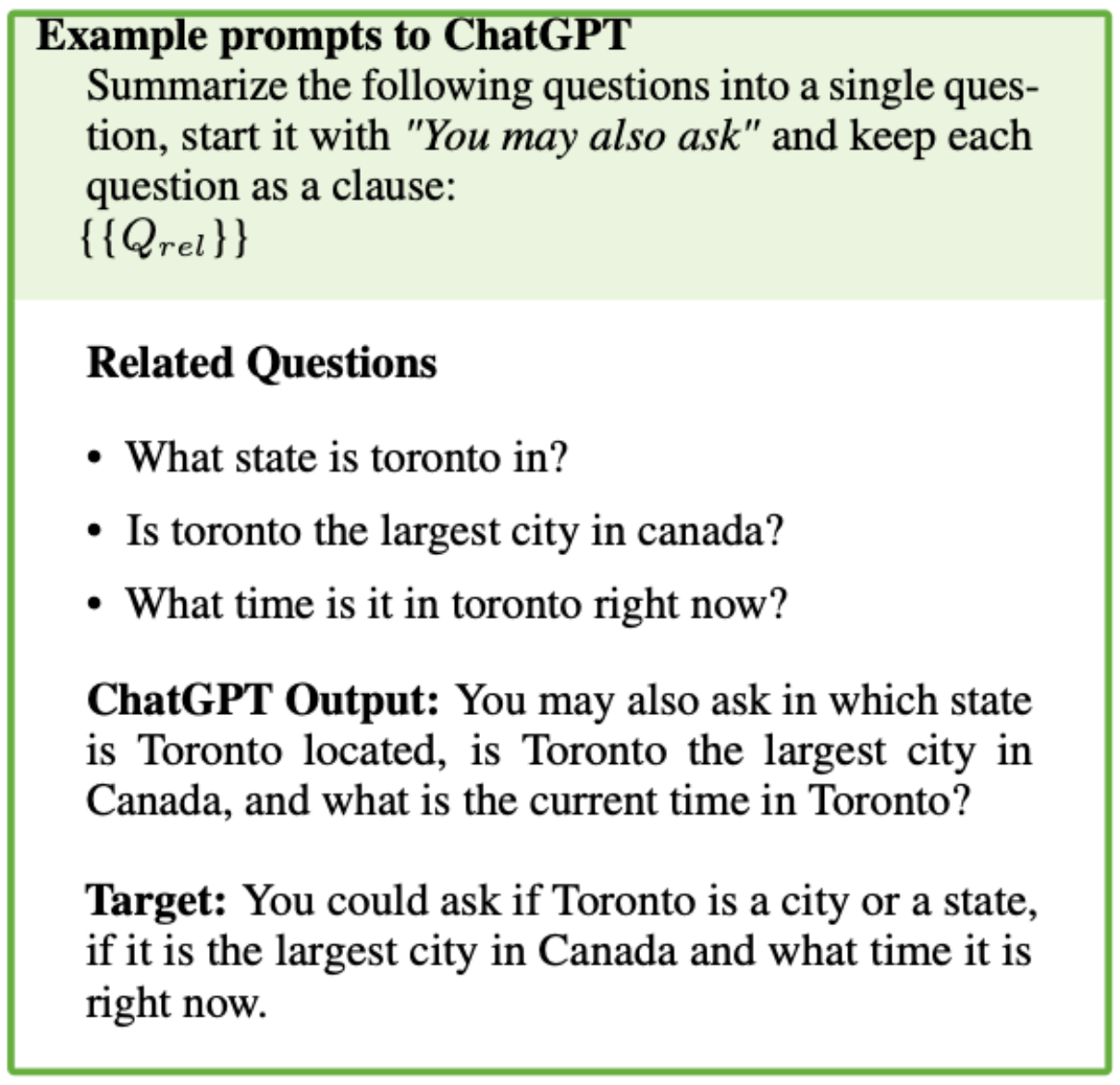}
\caption{The input prompt for ChatGPT, along with an example output and human ground truth (target).}
\label{fig:chatgpt}
\end{figure}

We prompted ChatGPT with the input related questions $Q_{rel}$, and asked to generate the hint using the prompt show in \Cref{fig:chatgpt}.

We find that ChatGPT in a zero-shot setting has significantly worse performance in terms of BLEU and ROUGE metrics, achieving the following performance on automated metrics shown in \Cref{tab:chat_gpt}.

\begin{table}[ht!]
    \centering
    \resizebox{1.0\columnwidth}{!}{
    \begin{tabular}{c c c c c c c}
    \toprule
     BLEU1 & BLEU2 & ROUGE1 & ROUGE2 & \\
    \midrule
     0.508 ($\blacktriangledown$ 12.2\%) & 0.297 ($\blacktriangledown$ 23\%) & 0.652 ($\blacktriangledown$ 9\%) &  0.403 ($\blacktriangledown$ 13.6\%)\\
    \bottomrule
    \end{tabular}}
    \caption{ChatGPT zero-shot performance on the task of hint generation. The relative difference w.r.t \ptg-BART is shown in parentheses.}
    \label{tab:chat_gpt}
\end{table}

Finally, while ChatGPT has reasonable performance in zero-shot settings, there are limitations in terms of fine-tuning such LLMs. First, models like ChatGPT are not scientifically reported and the model is not publicly available. Second, the sheer size of the model makes it impractical and impossible to use in voice assistants, where such hints are generated in real-time based on the user's questions, requiring the models to meet very strict latency requirements where the hint must be generated in less than 150 milliseconds.

\begin{table*}[ht!]
    \centering     
    \resizebox{1.0\textwidth}{!}{
    \begin{tabular}{p{5cm} p{2.2cm} p{10cm}}
    \toprule
    input & model &  hint\\
    \midrule
        \multirow{4}{5cm}{how many times can i enter wrong passcode on iphone? \texttt{[SEP]} can i unlock my iphone even if i am still paying for it? \texttt{[SEP]} why does messages on iphone 8 show half moon?} & \dhg-Bart & You may be interested to know how many times you can enter wrong passcode on iPhone, \textcolor{red}{or if you still paying for it}  and why messages on iPhone 8 show half moon. \\
        & \ptg-BART & You may want to know how many times you can enter wrong passcode on iPhone, \textcolor{blue}{or if you can unlock it even if you are still paying for it}, or why messages on iPhone 8 show half moon.\\
        & \rsb & You could ask how many times can i enter wrong passcode on iphone, or if i can unlock my iphone even if i still paying for it, or why messages on iphone 8 show half moon \\
        & \tb & You could ask how many times can i enter wrong passcode on iphone, or can i unlock my iphone even if i still paying for it, or why does messages on iphone 8 show half moon? \\
        \midrule
        \multirow{4}{5cm}{What is the largest horse that is alive? \texttt{[SEP]} Where does the word horse come from? \texttt{[SEP]} What is the collective name for a group of horses?} & 
        \tb & You can ask what is the largest horse that is alive, or where does the word horse come from, or what is the collective name for a group of horses?\\
        &\rsb & You might be interested to know what is the largest horse that is alive, or where the word horse comes from, or what is the collective name for a group of horses\\
        & \dhg-BART & You may want to know what is the largest horse that is alive, where \textcolor{red}{it comes from} and what is the collective name for a group of horses.\\
        & \ptg-BART & You may want to know what is the largest horse that is alive, where the \textcolor{blue}{word horse} comes from and what is its collective name for a group of horses.  \\
        \midrule
        
        \multirow{4}{5cm}{Who is the mother of cristiano ronaldo's twin's child? \texttt{[SEP]} Who is cristiano ronaldo's real wife? \texttt{[SEP]} How much money does earn cristiano ronaldo?} & DHG-BART & You may want to know who is the mother of Cristiano Ronaldo's twin's child, or who is his real wife. \\
        &\ptg-BART&You may want to know who is the mother of Cristiano Ronaldo's twin's child, or who is his real wife, or \textcolor{blue}{how much money he earns.}\\
        &\rsb&you might also be interested to know who is the mother of cristiano ronaldo's twin's child, or who is cristiano ronaldo's real wife, or how much money cristiano ronaldo earns\\
        &\tb& You can ask who is the mother of cristiano ronaldo's twin's child, or who is cristiano ronaldo's real wife, or how much money does earn cristiano ronaldo? \\
        \bottomrule
    \end{tabular}}
    \caption{Example hints generated by each model.}
\label{tab:sentences_examples}
\end{table*}